\pdfoutput=1
\documentclass[a4paper,twoside,10pt]{article}
\usepackage{a4wide,graphicx,fancyhdr,amsmath,amssymb,wrapfig,tabularx,amssymb}
\usepackage{float}
\title{\vspace{-30pt} Rank Ordered Autoencoders}

\author{
  Paul Bertens\thanks{Github: {https://github.com/paulbertens}; E-mail: {p.m.w.a.bertens@gmail.com}} \\
}
\date{\today}

\begin{document}

\maketitle

\begin{abstract}
A new method for the unsupervised learning of sparse representations using autoencoders is proposed and implemented by ordering the output of the hidden units by their activation value and progressively reconstructing the input in this order. This can be done efficiently in parallel with the use of cumulative sums and sorting only slightly increasing the computational costs. Minimizing the difference of this progressive reconstruction with respect to the input can be seen as minimizing the number of active output units required for the reconstruction of the input. The model thus learns to reconstruct optimally using the least number of active output units. This leads to high sparsity without the need for extra hyperparameters, the amount of sparsity is instead implicitly learned by minimizing this progressive reconstruction error. Results of the trained model are given for patches of the CIFAR10 dataset, showing rapid convergence of features and extremely sparse output activations while maintaining a minimal reconstruction error and showing extreme robustness to overfitting. Additionally the reconstruction as function of number of active units is presented which shows the autoencoder learns a rank order code over the input where the highest ranked units correspond to the highest decrease in reconstruction error. 
\end{abstract}

\section{Introduction} 

\subsection{Autoencoders} 
Autoencoders are a common technique used for the unsupervised learning of representations of the underlying data\cite{Le2011, Wang2014, Baldi2012, Rifai2011, Ranzato2007, Erhan2010}. Based on artificial neural networks they have as output target the given input, and must learn to reconstruct the input through some representation in the hidden layers. While standard autoencoders are effective at learning a representation that is smaller in dimensionality than the input, when the representation is over-complete, i.e. there are more hidden units than input units, a standard autoencoder can learn the identity function instead of meaningful features. To avoid this problem many techniques have been developed, most notably sparse autoencoders, which put an extra cost on the hidden layer to create sparse representation\cite{Coates2011, Kavukcuoglu2010} or restrict the number of allowed non-zero activations\cite{Makhzani2014} and denoising autoencoders which add noise to the input and try to reconstruct the original input from this noisy input\cite{Vincent2008, Zhou2012, Vincent2010}. These techniques require the autoencoder to learn more meaningful features that can describe the data instead of just copying the current input. 

\subsection{Sparse coding} 
There have been several attempts at achieving sparse over-complete representations as they have many desirable properties over conventional dense codes\cite{Golomb1990, HonglakLeeAlexisBattleRajatRaina2006, Olshausen2004, Olshausen1997}. A dense code is far more difficult to decode, as more units are required for decoding and each unit has less individual meaning. The code is also unlikely to be linearly separable which is an important property in learning representations. Dense codes have the problem of catastrophic forgetting as well, as all units get updated for each input, while in a sparse code only a small subset gets updated. Furthermore a sparse code allows for more efficient computation as less input units are required to compute the output. Still a too sparse code is also not optimal, as it results in less possible ways to encode the input and reduces the similarity between the representation and the inputs. Learning similarity between inputs is needed to not have to relearn many common features between them and to be able to compare inputs from their representation. One thus wants to strike a good balance between these such that each input is represented by a sparse subset of features, usually in the context of autoencoders this is done by setting a sparsity penalty on the hidden unit representation, this however adds an extra hyper-parameter to be learned and creates a trade-off between sparsity and reconstruction error. An alternative is to select only the top $k$ units for activation as was done in the k-sparse autoencoder paper\cite{Makhzani2014} but this suffers from a similar limitation as we have to specify a value for the hyperparameter $k$. Alternatively one could train on a small set of units that is incrementally updated which is what is done in the incremental autoencoder approach\cite{Zhou2012}, where a small set of units is trained first and new units are added while also merging existing units. This however is an incremental approach and the units are not directly trainable in parallel and again adds extra hyperparameters for merging and adding. 

\subsection{Ordered codes}
Another related approach is based on learning ordered codes, one such example is the use of nested dropout\cite{Rippel2014}, which is an extension of standard dropout\cite{Srivastava2014}. Where instead of dropping 50\% of the hidden units at random, units are dropped in order such that units can depend on units before them while in standard drop-out this is not the case and each unit has to be mostly independent. This is done by selecting an index according to some exponential distribution and dropping all units after the index. In the paper one issue was that later units had the problem of being very likely to be dropped and therefore were difficult to train, to avoid this they fixed each of the units in order after they had been trained sufficiently. However this significantly increases training time and due to the fixed distribution on the output only strictly ordered codes can be learned which limits the number of possible ways to represent the input. 

\subsection{Objective} 
Some of the previous limitations of learning meaningful features in autoencoders is thus that they either require sequential computation that is difficult to parallellize, need some type of stochastic noise, or require extra hyperparameters for the sparsity level. Therefore the desired objective of this paper is to find the optimal weights of an autoencoder such that it is capable of reconstructing using a minimal number of output units without the need for a sparsity parameter or added noise and still remain easily parallelizable. We thus want to minimize the $L_0$ norm of the output units on the condition that the reconstruction error stays minimal.
To formalize this means we want to minimize the following function:\\
\begin{equation} \label{eq1}
\begin{split}
E &=||\bold{x}-\bold{\hat{x}}||_2 + \lambda ||\bold{y}||_0 \\
 &=||\bold{x}- f(\bold{xW})\bold{W^T})||_2 + \lambda ||f(\bold{xW})||_0
\end{split}
\end{equation}

\begin{figure}[H]
\centering
\includegraphics[scale=0.5, width=0.6\textwidth]{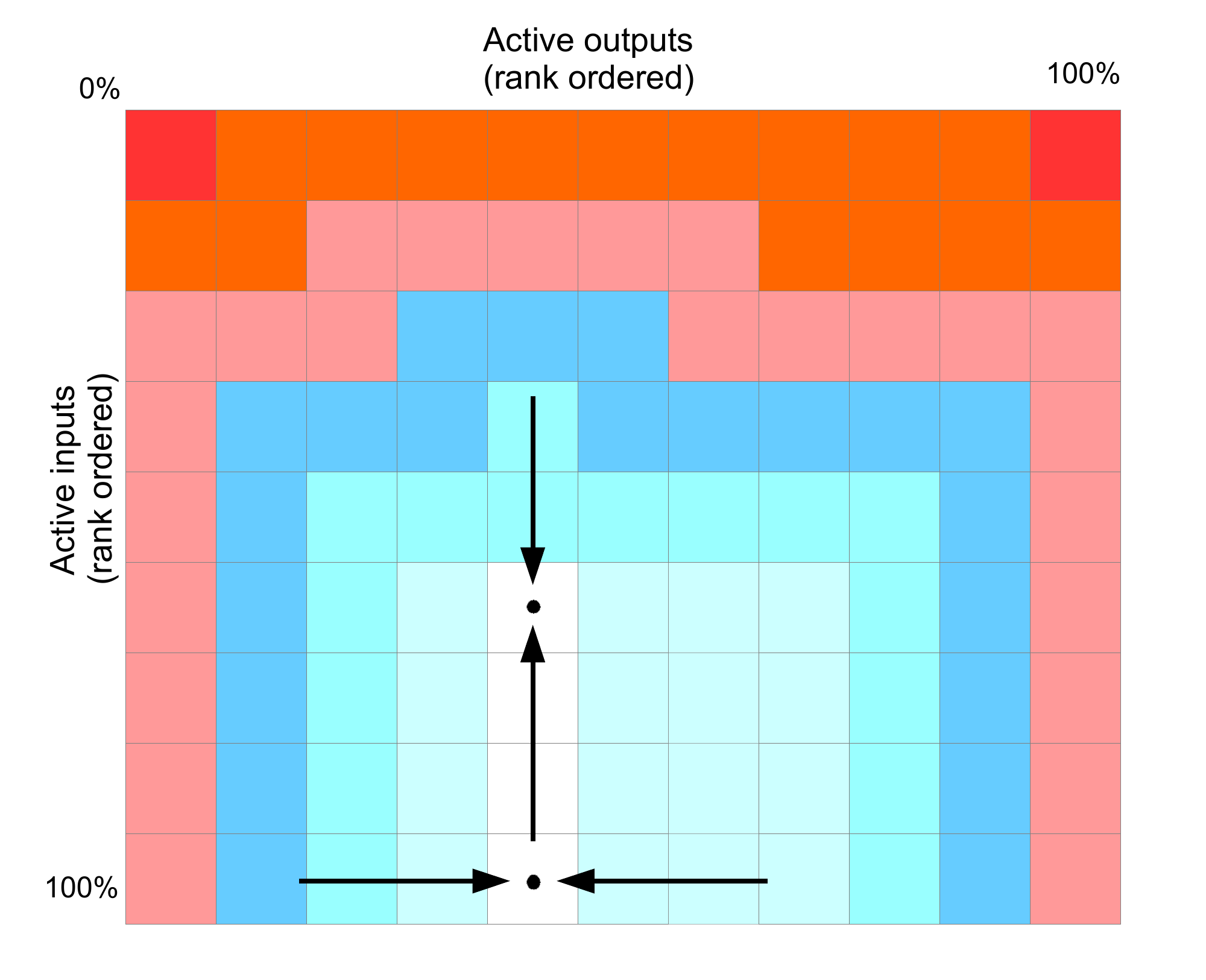}
\caption{An illustration of a hypothetical reconstruction error surface as a function of number of active outputs and inputs, darker regions indicate higher errors and the units are sorted by their output value. In case of minimizing the number of active output units the progressive reconstruction causes all units to try and minimize the reconstruction error by moving horizontally, optimizing the number of units required to reconstruct. (We could also theoretically move vertically with a reverse rank ordered autoencoder and minimize the number of active input units required for reconstructing.)}
\label{reconstructionsurface}
\end{figure}

\begin{figure}[H]
\centering
\includegraphics[scale=0.5, width=0.8\textwidth]{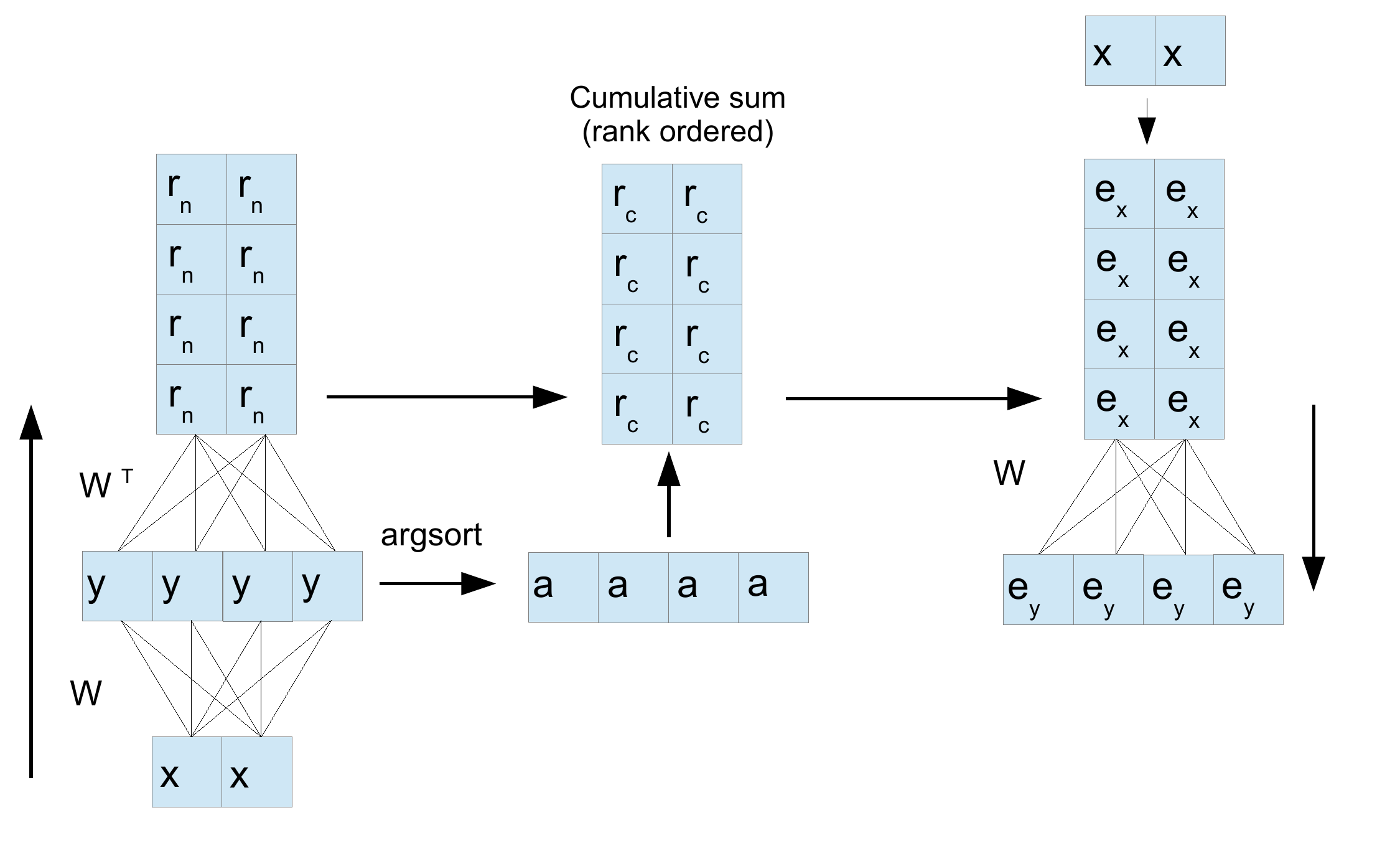}
\caption{The proposed approach for minimizing the number of outputs required to reconstruct the input. $x$ is the input, $y$ the hidden representation, $r_n$ the normal individual reconstructions, $a$ the indexes that would sort $y$, $r_c$ the progressive reconstruction, $e_x$ the error with respect to the real input and $e_y$ the back-propagated error at the hidden units.}
\label{network_diagram}
\end{figure}

This objective function is not convex and not directly differentiable due to the $L_0$ norm and the unknown $\lambda$, in practice usually this objective function is changed to minimizing the $L_1$ norm and setting $\lambda$ as an extra hyper parameter\cite{Kavukcuoglu2010, Candes2005}. However we take a different approach, in order to try and solve this minimization problem a method is proposed and implemented based on ordering the output representation of a standard autoencoder by their output values and reconstructing from this representation progressively using a cumulative sum (also known as prefix sum, i.e. $y_{i}=\sum_{k=0}^{i}{x_k}$). We would thus like to minimize the following function instead:\\
\begin{equation} \label{eq2}
\begin{split}
E &=\sum_{j=0}^m \sum_{k=0}^j||\bold{x}-\bold{\hat{x}}^{ro}_{m-k}||_2 \\
&=\sum_{j=0}^m \sum_{k=0}^j||\bold{x}-\sum_{l=0}^{m-k} \bold{\ddot{x}}^{ro}_{l}||_2
\end{split}
\end{equation}
Where $m$ is the number of output units, and $\bold{\hat{x}}^{ro}_{m-k}$ the progressive reconstruction using a cumulative sum over the rank ordered individual reconstructions $\bold{\ddot{x}}^{ro}_{l}$ (ordered by the rank of output $\bold{y}$ from high to low). In this ordered cumulative sum domain the function does become directly optimizable (see also figure \ref{reconstructionsurface} and \ref{network_diagram}). The $\sum_{k=0}^j||\bold{x}-\bold{\hat{x}}^{ro}_{m-k}||_2$ term intuitively means that we would like to have the lowest reconstruction error left for low ranked units, and the highest error for high ranked units. If we take some arbitrary reconstruction error $\epsilon_k$ at unit $k$ we will have that $\sum_{j=0}^{m} \sum_{k=0}^j \epsilon_{m-k} \geq \sum_{j=0}^{m} \sum_{k=0}^{j} \epsilon_{m-k-1}$. This implies that units with a high ranking should learn to minimize the error as much as possible to ensure that the total error is as low as possible. If an error is left the sum of the cumulative sum of this error always increases proportional to the number of units that come after it (we take the reverse cumulative sum $m-k$), meaning the total error always increases. This means for example that the error vector [0.5, 0.2, 0.1] has an higher error than [0.5, 0.3, 0] (if we were to take simply the sum this would not be the case). Minimizing this function will therefore ensure that units with a high output (high ranking) will reconstruct most of the signal while successive units will reconstruct less and less. We not only implicitly minimize the ordered $L_0$ norm this way, i.e. $\sum_{j=0}^m \sum_{k=0}^j sign(y^{ro}_{m-k})$, we also minimize the sum of the cumulative sum of the remaining non-zero components, $\sum_{j=0}^m \sum_{k=0}^j y^{ro}_{m-k}$. In order to reconstruct a large part of the signal (and maximally reduce the error) the weights have to be large enough, this in turn means the output will increase, thus we implicitly learn to also increase the output of the high ranked units. It also means as we maximally decrease the error in high ranked units, units of lower rank will either have to be zero or have small weights which will decrease the output of the lower ranked units. So we can indirectly solve the actual desired objective function:
\begin{equation} \label{eq3}
\begin{split}
 E &= ||\bold{x}-\bold{\hat{x}})||_2 + \lambda \sum_{j=0}^m \sum_{k=0}^j y^{ro}_{m-k} \\
 &\leq ||\bold{x}-\bold{\hat{x}})||_2 + \lambda \sum_{j=0}^m \sum_{k=0}^j sign(y^{ro}_{m-k}) \\
  &= ||\bold{x}-\bold{\hat{x}})||_2 + \lambda ||\bold{y}||_0 + \sum_{j=0}^{m-1} \sum_{k=0}^{j} sign(y^{ro}_{m-k})
 \end{split}
\end{equation}
by minimizing the progressive reconstruction instead (eq. \ref{eq2}) without having to know the sparsity parameter $\lambda$. The difference between eq. \ref{eq1} and eq. \ref{eq3} is only that instead of minimizing the $L_0$ norm we indirectly minimize a type of ordered $L_0$ norm. If we have that $0\leq y_j \leq 1$ then $\sum_{j=0}^m y^{ro}_j \leq \sum_{j=0}^m sign(y^{ro}_j) = \sum_{j=0}^m sign(y_j) = ||\bold{y}||_0$. Intuitively because of the property that $\sum_{j=0}^m \sum_{k=0}^j y^{ro}_{m-k} \geq \sum_{j=0}^m \sum_{k=0}^j y^{ro}_{m-k-1}$, the minimum solution of eq. \ref{eq3} is the one has as many zeros as possible for lower ranked units, and should move as much of the output to the higher ranked units. This is identical to what eq. \ref{eq2} is doing, meaning we can indeed indirectly solve eq. \ref{eq3} by minimizing eq. \ref{eq2}. It is also clear that there is a relation between eq. \ref{eq1} and eq. \ref{eq3} through the $L_0$ norm term (taking the $L_0$ norm of any permuted vector, including the rank ordered vector, is the same as the $L_0$ norm of the original vector), while an exact mathematical derivation of whether the solutions of minimizing these equations are identical is outside the scope of this paper, we do show experimentally that we indeed achieve the expected sparsity on the output. 

In the approach section we expand further upon this idea giving both algorithmic pseudo-code and justifications for the used technique. First the basic approach of minimizing the number of outputs required for reconstruction is presented. A custom derivative function is then also derived to help minimize the given objective function (eq. \ref{eq2}). Additionally details are given on algorithmic complexity and we expand upon the relation to rank order coding, PCA(principal component analysis) and nested dropout. In the experiment section the model is tested on patches of the CIFAR10 dataset by looking at learned features, reconstruction errors, convergence rates and sparsity levels. Finally we conclude with a summary of the results and present possible future extensions and improvements to the model.

\section{Approach} 

\subsection{Rank Ordered Autoencoders} 
In order to minimize the number of required outputs for reconstruction we can progressively, in order, reconstruct the input by the rank order of the output values. We can then compute the error with respect to this progressive reconstruction and minimize this error for each of the output units by using standard stochastic gradient descent and back-propagation (see also figure \ref{network_diagram}). Each unit thus tries to minimize the remaining reconstruction error left by previous units that are higher ranked, and we do this in parallel through the use of a cumulative sum. If this error becomes zero the error at the remaining units will also become zero resulting in the use of minimal output units for reconstruction. This way we can solve the problem of finding the set of weights that minimizes eq. \ref{eq2} and also indirectly eq. \ref{eq3}. In the classical approach we would calculate the sum of the weights multiplied by the output values when doing the reconstruction, in this case we calculate the cumulative sum of the reconstruction ordered by their output ranks. This cumulative sum is capable of assigning an unique error to each unit dependent on their rank, while normal summation would 'blur' this error. \\

\noindent To illustrate the process (numpy like notation for convenience\footnote{'[]' indicate indexing, ':' is selecting all rows/cols at that location, and '[:, args]' is indexing the columns by the args vector. Broadcasting means if we have a matrix of shape (n,m) and vector of shape (n) we apply the operation to all columns of the matrix, and thus get an output of (n,m).}):\\
1: calculate forward pass; $\bold{y}=f(\bold{xW})$ (standard matrix-vector multiplication)\\
2: argsort output; $\bold{args} = argsort(\bold{y})$ (high to low)\\
3: reconstruct from each unit; $\bold{r}=\bold{W}*\bold{y}$ (broadcasted multiplication)\\
4: reconstruct progressively in order; $\bold{r}[:, \bold{args}] = f(cumsum(\bold{r}[:, \bold{args}], axis=1))$\\
5: calculate error; $\bold{e_x}=\bold{r}-\bold{x}$ (broadcasted subtraction)\\
6: backprop error; $\bold{e_y}=sum(\bold{W}\cdot \bold{e_x}, axis=0)$ (elementwise multiplication and sum)\\
7: multiply by derivative; $\bold{e_y}=\bold{e_y}\cdot f'(\bold{y}, \bold{e_y})$\\
8: calculate gradient input; $\bold{g_x}=\bold{e_x}*\bold{y}$ (broadcasted multiplication)\\
9: calculate gradient output; $\bold{g_y}=\bold{x}\bold{e_y}$ (outer product)\\ 
10: update weights; $\bold{W} =\bold{W}+\epsilon(\bold{g_x}+\bold{g_y})$\\

\begin{figure}[h!]
\centering
\includegraphics[scale=0.5, width=0.49\textwidth]{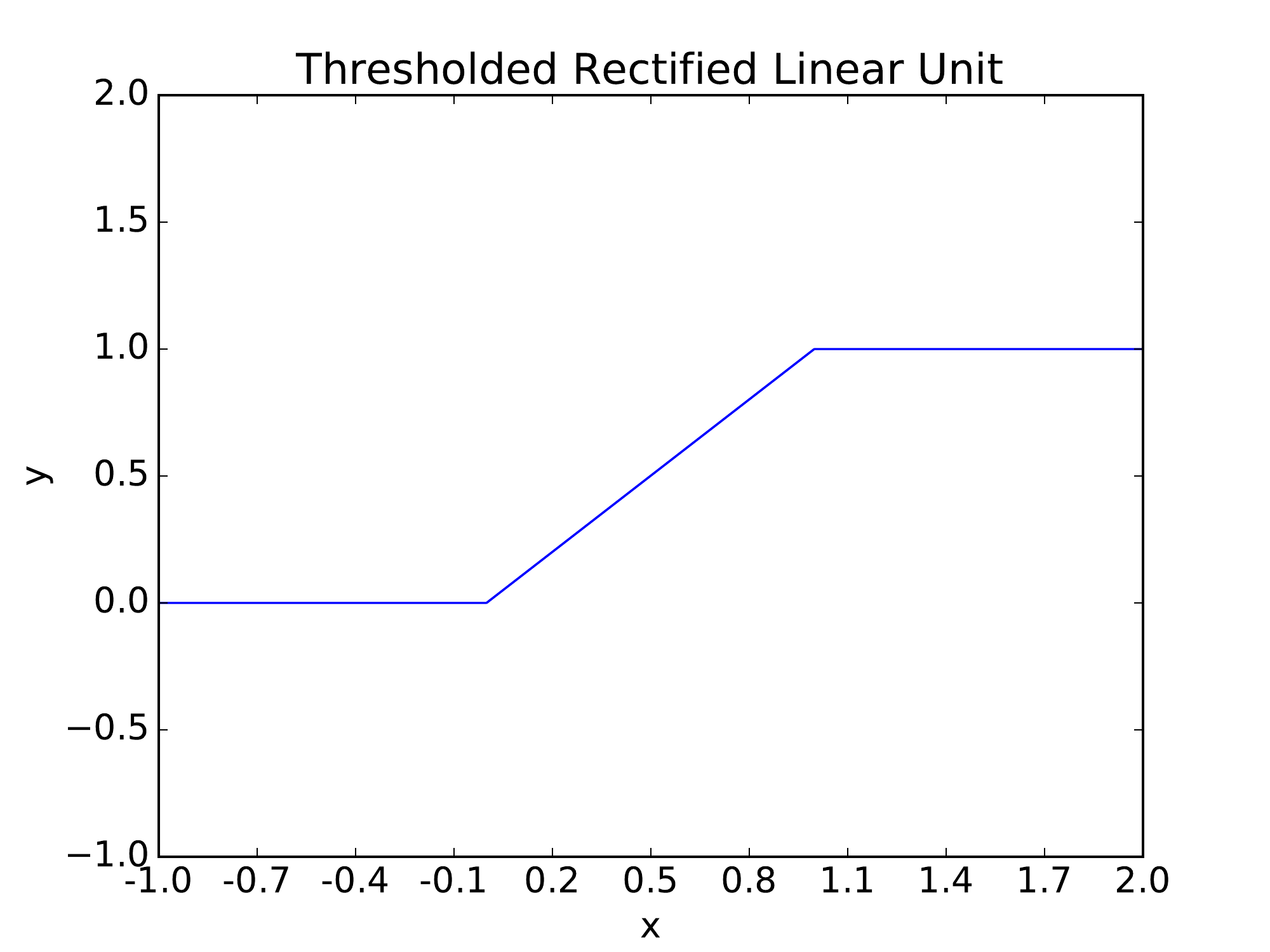}
\includegraphics[scale=0.5, width=0.49\textwidth]{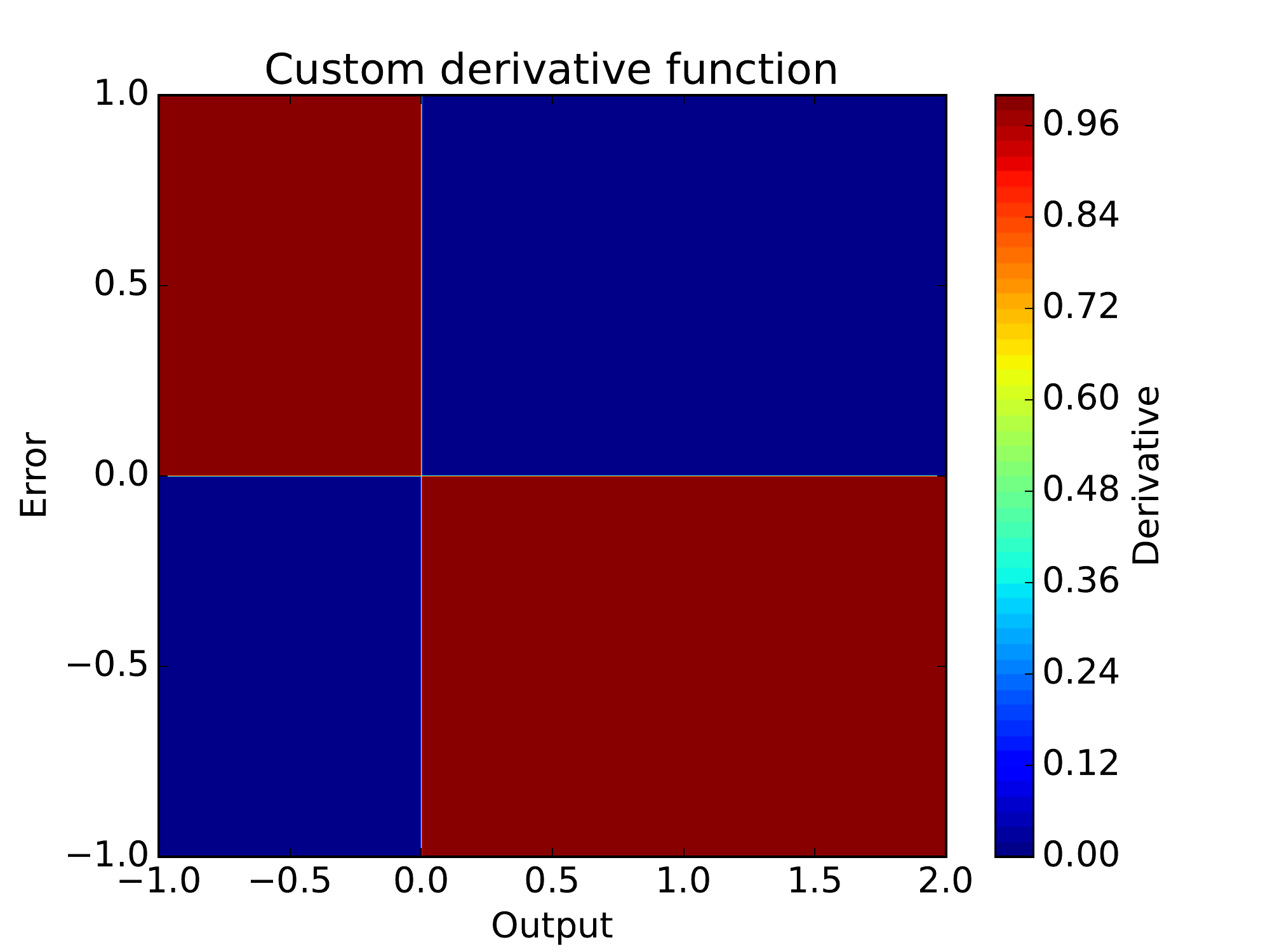}
\caption{The activation function used $f=min(1, max(0, x))$ (left) and the custom derivative of the activation function $f'=(sign(error) \neq sign(output))$ (right), which is dependent on both the error and output. This function causes hidden units to only be able to move towards 0 to both avoid dead units and minimize the number of active units. All other changes to the output is determined solely by the weight updates of the error at the input.}
\label{activation}
\end{figure}

Instead of the standard rectified linear unit, the output is thresholded to 1 and the new activation function becomes $f=min(1, max(0, y))$ (TReLU), see figure \ref{activation}. This is done to avoid potentially blowing up the output activation which would blow up the gradient and it is also a desirable property for potentially deeper networks to avoid exponentially increasing output. However since the summation is highly likely to exceed 1 we divide the output by the number of input units, if we assume each input unit is between 0 and 1 and each weight is approximately between -1 and 1 (in case of tied weights this is likely the case as reconstruction weights are close to actual input values) then the output will never actually exceed 1.  

A different derivative is also used for the activation function (see figure \ref{activation}), instead of the real derivative, which is zero when $y<0$ and $y>1$, we use a maybe less mathematical rigorous but more sensible version that can avoid dead units. The standard derivative often leads to inactive dead units as the derivative is always zero for negative outputs so instead we derive the derivative from the movement of the units that we desire. We want to move inactive units up if required, this is the case when the error at that unit is still high meaning previous higher ranked units did not properly reconstruct and we should 'recruit' this unit. We also want to move active units down if required, meaning too many units are active and the reconstruction error is too high. Moving negative units more down wouldn't make sense as these were already zero and this would not improve the objective function and instead would merely destroy previously learned representations, similarly moving positive units more up is not desired as this is already done by the gradient calculated at the input. This thus reduces to the following desired derivative: $f'=(sign(error) \neq sign(output))$. There is also no bias on the hidden units, instead the bias is assumed to be implicitly learned from the input. This is done because if a unit has to change its activation value it should be as a function of other units, which is implicitly contained within the input values, instead of as a function of a bias value.   

As was mentioned in the introduction this approach of reconstruction and weight updating can be seen as a way of minimizing the ordered $L_0$ norm of the output representation while keeping the reconstruction error minimal. This is in stark contrast to the usual approach of learning sparse representations that set specific sparsity target values which try balance reconstruction with sparsity. In our case it is assumed that some units will actually increase  reconstruction errors, as they likely represent features of different inputs adding them to the reconstruction increases the error, but too few units also increase the error as we have less units to represent the input (see figure \ref{reconstructionsurface}). Thus we try to minimize the number of units while keeping the reconstruction error minimal. If the error increases the proposed update rule implicitly adds new units to be able to fix this error, while if there are too many units that increase the error the update rule will reduce this error by removing these units from the current reconstruction, this is achieved by the custom derivative function (Figure \ref{activation}) and the rank ordered progressive reconstruction (Figure \ref{network_diagram}).

\subsection{Algorithmic complexity} 
The algorithmic complexity does not actually increase compared to standard auto-encoders. Since the forward pass is left unchanged the only additions are a sort on the outputs and a cumulative sum on the reconstruction. Let $n$ be the number of input units, $m$ be the number of output units, $x$ the input, $w$ the weight matrix and $y_j$ the output at unit $j$. Then the forward pass which is a matrix vector product becomes $y_j=\sum_{i=0}^n{x_i \cdot w_{ij}}$ which is simply $O(n\cdot m)$. It is known that sorting of $m$ values can be efficiently done in $O(m \log(m))$ time and since $O(m\log(m)) \leq O(n\cdot m)$ for $n\geq log(m)$ (which is highly likely as $\log_2$ of e.g. 1024 hidden units is only 10) our complexity remains $O(n\cdot m)$. The reconstruction complexity using the cumulative sum, $r_{j, i}=r_{j-1, i}+y_j \cdot w_{ji}$, is $O(n \cdot m)$ which is the same as when taking the normal sum. Other computations are always standard element-wise products or summations never exceeding $O(n\cdot m)$ and thus overall algorithmic complexity remains unchanged. Memory complexity also remains the same, we do require to keep the cumulative sum however this does not increase the standard memory complexity as it is bound by the weight matrix $O(n \cdot m)$. For input vectors and output vectors the cumulative sum requires an extra $m$ outputs resulting in a memory complexity of $O(n \cdot m)$, however it should be noted this only holds for input vectors, for input matrices when memory is no longer bounded by the weight matrix but by the input size, memory complexity does increase by a factor proportional to the matrix height (i.e. mini-batch size).

\subsection{Computational efficiency using ordered sparsity} 
The induced sparsity combined with the ordered representation can further be exploited to increase computational efficiency and reduce memory requirements. As was mentioned before we are computing the cumulative sum over the output units which in the case of vectors does not increase the memory bound however when scaling up to using mini-batches this significantly increases memory requirements over standard matrix-matrix multiplications (increased by mini-batch size $d$). This increase in memory can somewhat be reduced by only reconstructing from the active units, if we assume only $k$ active units, the increase in memory is only by a factor of $k$ and so instead of a memory complexity of $O(n \cdot d \cdot m )$ we get a complexity of $O(n \cdot d \cdot k)$, since we learn extremely sparse representations this number can be much lower than the actual number of output units bringing it closer to the standard memory bound of $O(n \cdot d)$. Further more in potentially deeper architectures the ordered sparse output can be used to compute the forward pass on only these active units resulting, in case of using vectors, in a forward pass of $O(n \cdot k)$ instead of $O(n \cdot m)$. This allows for a very large number of hidden units while still keeping computational cost low. 

\subsection{Relation to PCA}
If we look at this approach in a sequential way instead of the presented parallel approach it becomes easier to see the relation to PCA (principal component analysis). The relation between standard autoencoders and PCA is well known however ranked autoencoders make this relation much more clear. The highest ranked unit $j$ gets reconstructed first, lets call this reconstruction $\hat{x}_j$ which thus means we try to minimize $||x-\hat{x}_j||_2$ then we add to this reconstruction the second highest ranked unit $j+1$ which means we have to minimize $||(x-\hat{x}_j)-\hat{x}_{j+1}||_2$ and we do this for each non-zero output unit. We thus subtract each component from the input, and learn the next component on the remainder, this is equivalent to performing PCA. However in contrast to PCA we have a non-linear output, so we are doing a type of conditional non-linear PCA. The non-linearity (ReLU) results in the selection of a subset of components, those that exceed zero, and weights each of the components by their output value. This means we can learn a much larger representations than PCA and we actually get components that are dependent on the input and only apply to some subset of the data instead of components that apply to all of the data. This allows for easier separability between inputs as we have a different subset of components for each input.  

\subsection{Relation to dropout}
Another way to see this approach is as a deterministic version of nested dropout on the sorted output. Since we sort the output we can have an arbitrary output instead of requiring a strictly ordered representation as was the case in the original nested dropout paper\cite{Rippel2014}, allowing different orders at the output allows for more possible ways to encode the input. While in standard dropout we have $2^m$ possible sub-models\cite{Srivastava2014}, in the case of nested dropout we only have $m$ possible sub-models.  It is possible to compute all these $m$ models in parallel with little overhead in memory as there is an order in the sub-models such that each model is dependent on the previous model. Computing these sub-models in parallel can be achieved using a cumulative sum as proposed in this paper. This potentially also still provides the overfitting robustness provided by dropout and since we additionally try and learn a representation of the data that uses as few non-zero components as possible, a type of minimum description length\cite{Grunwald2005}, it should make the approach extremely robust to overfitting. 
\newpage
\subsection{Relation to neuroscience and rank order coding} 
The idea of learning rank ordered representations is not new, and there have been many attempts in the past especially in the field of neuroscience\cite{Gautrais1998, Delorme2001, Galluppi2011}. However biologically realistic functions and mechanisms are commonly used and the continious value is usually discarded keeping only the rank information. Iterative methods are then employed to compute the outputs which are very difficult to parallelize making it much less practical. They also do not necessarily try to explicitly minimize the reconstruction error but instead use STDP (Spike time dependent plasticity)\cite{Delorme2001} or similar like learning rules for learning the internal representation. This means that  reconstructing from this model cannot always be done directly so there is less guarantee on the amount of information maintained by the representation. 

The continuous value in the rank ordered autoencoder can actually be interpreted as the spike timing, instead of the classical assumption that it is the firing rate. High outputs 'fire' first and we get a progressive spike wave ordered by their spike timings, we then try to explicit learn to reconstruct from this ordered spike wave. Discarding the exact timing while robust to contrast variation also reduces the luminance information contained in these spike timings, so we can keep more information using continuous values instead of rank values with little increase in learning complexity. Spike timings are biologically more realistic than firing rates as the rapid processing of visual stimuli observed in the brain cannot be achieved with the limited firing rate of neurons\cite{Gautrais1998}. These type of rank order codes encode most of the information in the initial few spikes which allows for more rapid inference to be made when still only a few neurons have fired\cite{VanRullen2001}. This makes the rank ordered autoencoder biologically much more realistic than the standard feed-forward networks that do not take into account this order in the spike timings. 

\section{Experiment}
\subsection{Method}
In order to test the proposed approach we apply it to patches of the CIFAR10\footnote{https://www.cs.toronto.edu/~kriz/cifar.html} dataset which consists of a set of 50000 training and 10000 test samples of natural images. The implementation was made in python using the standard numpy library. Currently only a CPU version is implemented however the approach is easily parrallizable and can be made to run efficiently on a GPU, at test time the rank ordered autoencoder is actually identical to the standard one, only the training is modified. The experiments were run on a i5-2500k CPU and run for 60 epochs. Each epoch consists of a random 7x7 patch extracted from each image in the dataset (for the test set we extract 5). Importantly there was no preprocessing done on the data (usually standardization and whitening is performed). This was avoided to evaluate the performance on unprocessed data, and it is assumed the autoencoder will actually extract the mean from the data in its first component. 

The number of hidden units were set to 169, and weights between the encoder and decoder are tied. L2 norm clipping\cite{Pascanu2012} plus a learning rate was used for updating the weights. Norm clipping clips the norm of the gradient to not exceed a certain threshold, this allows for faster learning in the initial few epochs as the error is much higher for the higher ranked units. The initial norm clip was set to 0.1 and the learning rate to 1. The learning rate was reduced by 10\% if the training error did not improve by more than 1\% after an epoch, the norm clip was not changed. Only a training set and test set was used as no validation set was needed for tuning any hyper-parameters. The learning rate and norm clipping were tuned using only the training error. 

\subsection{Results} 

\begin{figure}[H]
\centering
\includegraphics[width=0.8\textwidth]{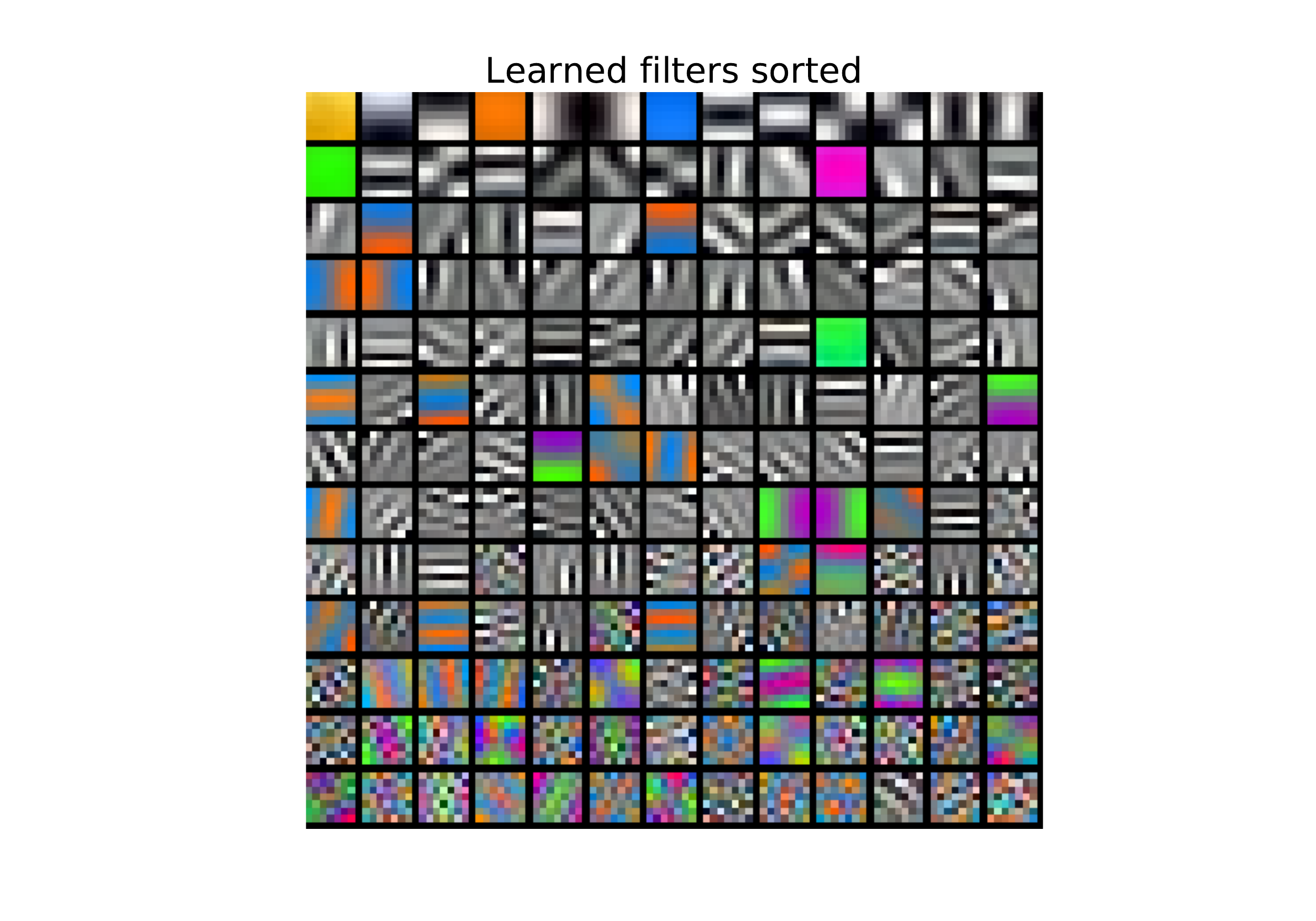}
\caption{The learned filters (best seen in color), normalized between 0 and 1 and sorted by their average activation. Showing learned edge filters and color patches that become increasingly complex.} 
\label{cifar_learned_filters}
\end{figure}

\begin{figure}[H]
\centering
\includegraphics[width=0.45\textwidth, height=150pt]{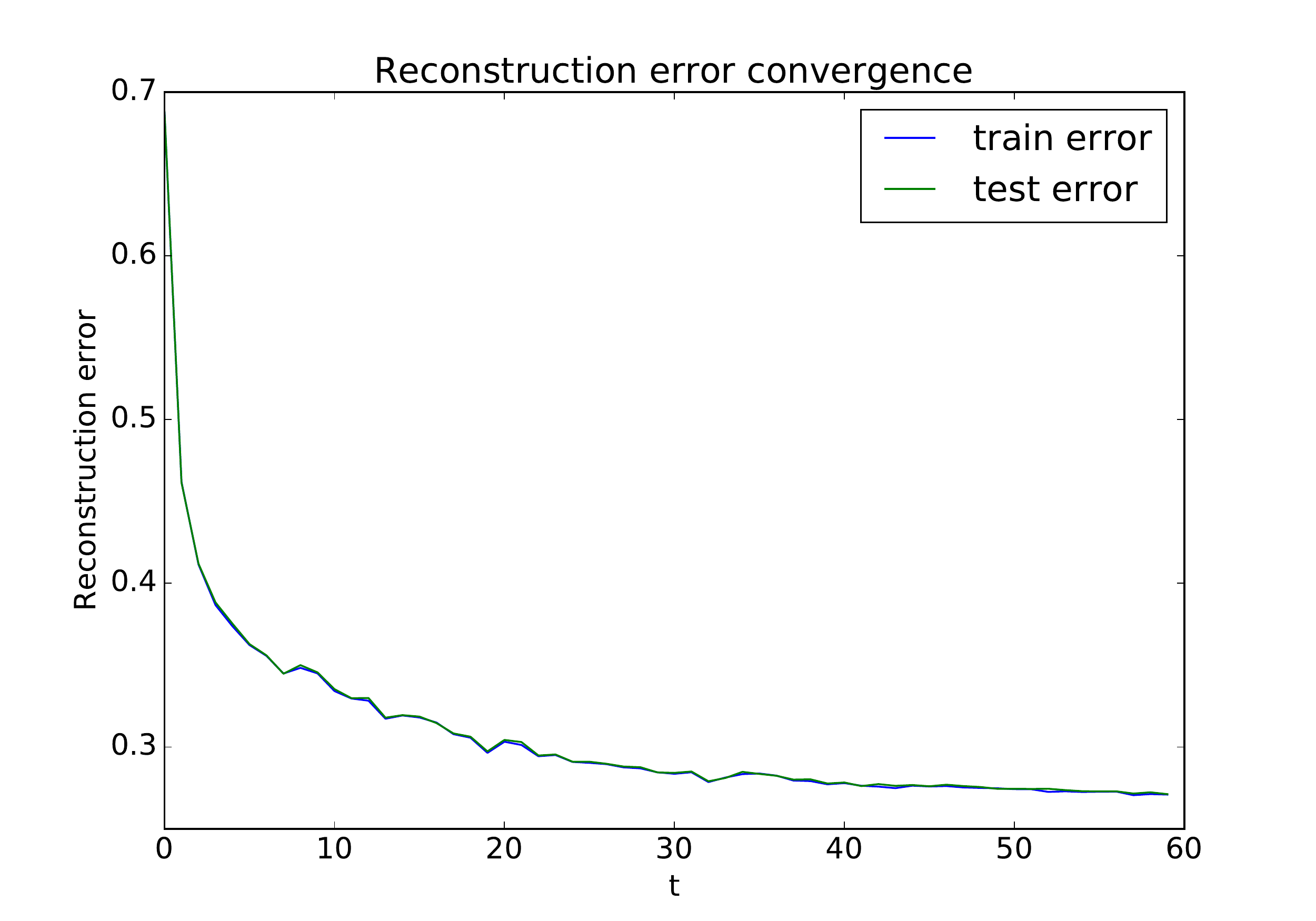}
\includegraphics[width=0.45\textwidth, height=150pt]{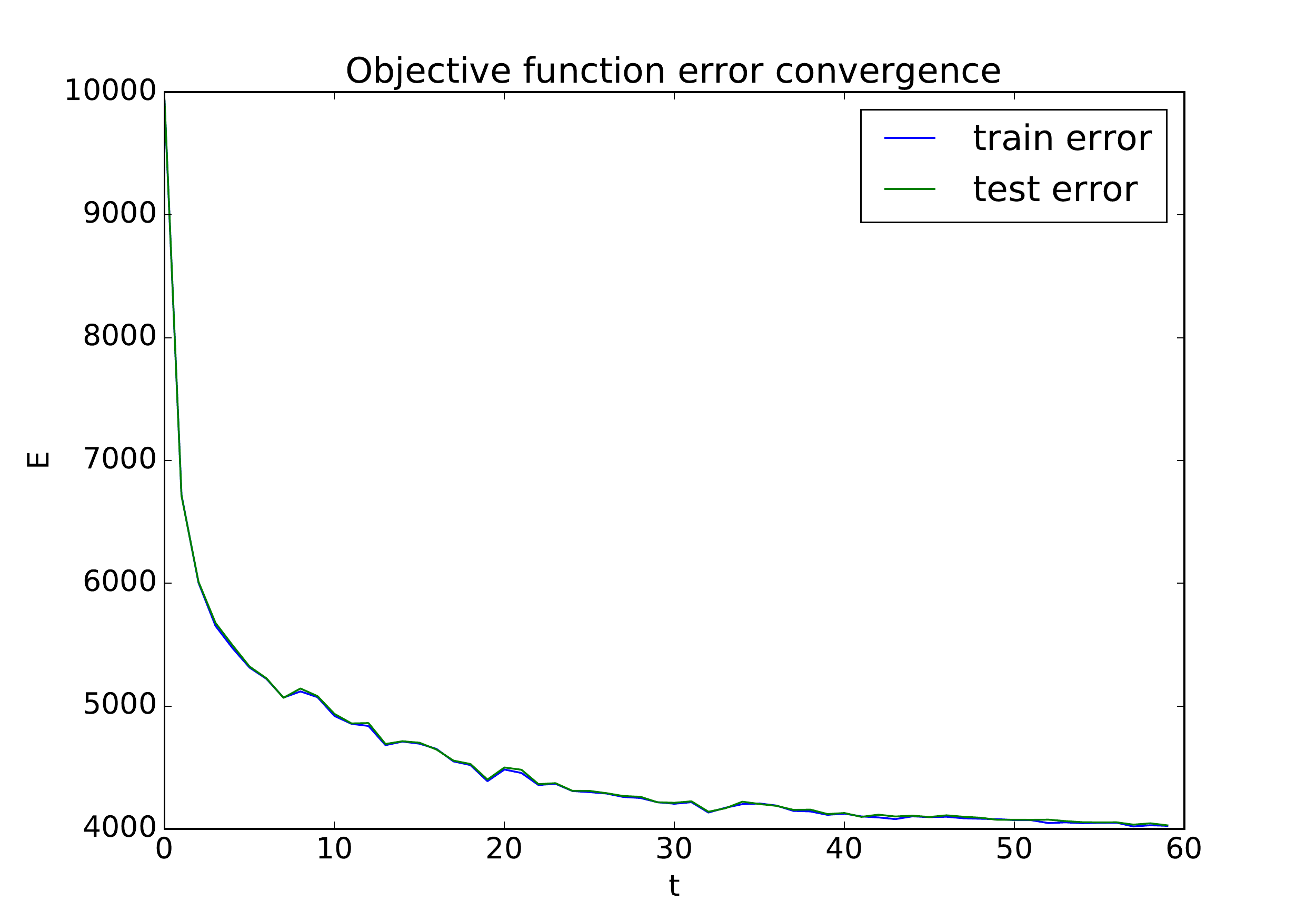}
\caption{Reconstruction error ($L_2$-norm)(left) over time, showing an initial rapid reduction in the error in just a few epochs stabilizing around 0.28,  and objective function error (eq. \ref{eq2})(right). There is notably no visible overfitting on the test set.} 
\label{cifar_reconstruction_errors}
\end{figure}

\begin{figure}[H]
\centering
\includegraphics[width=0.7\textwidth]{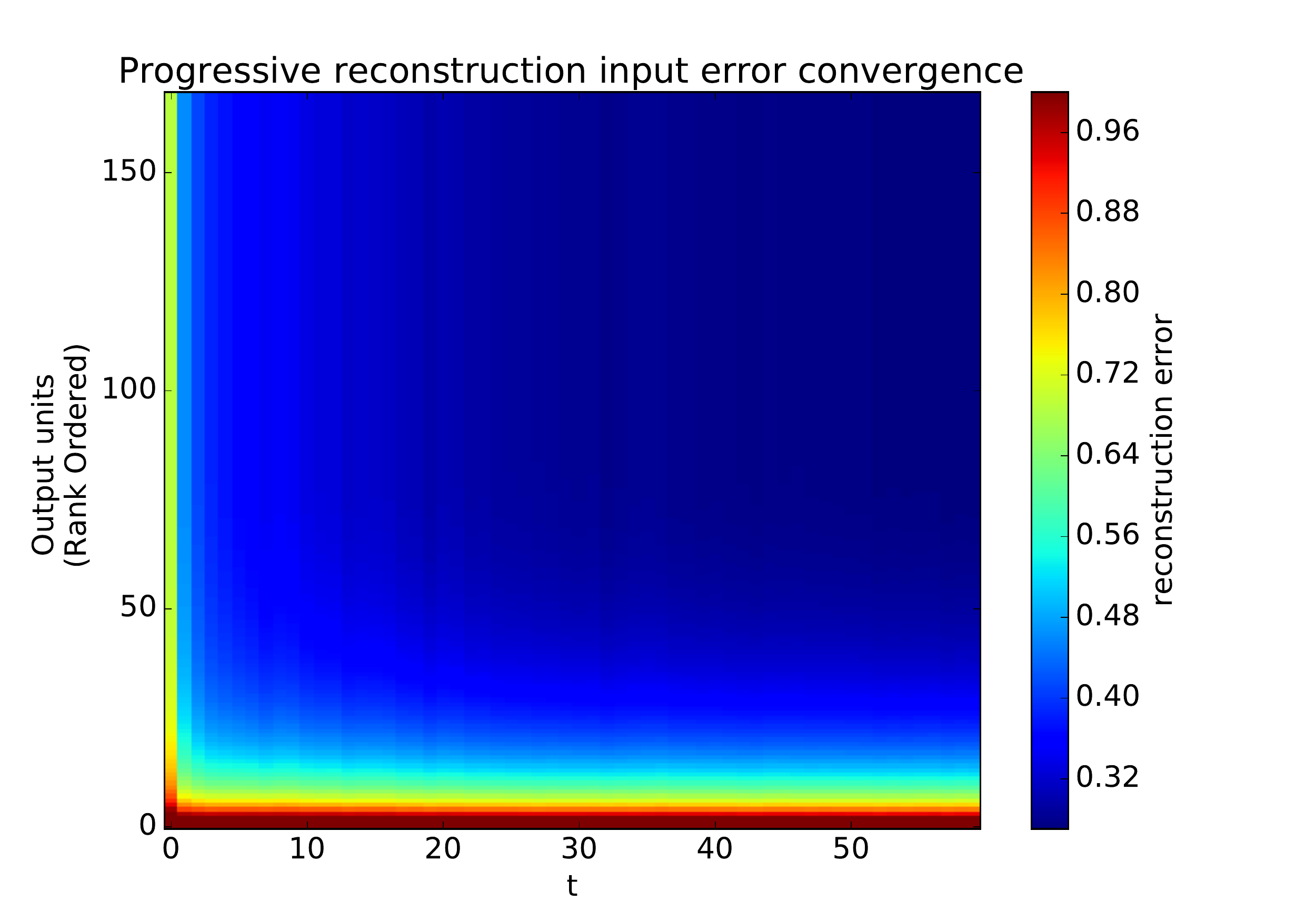}
\caption{The errors of the progressive reconstruction as a function of the sorted outputs over time. Showing an initial rapid reduction in error for the higher ranked units that gradually decreases for each successive epoch meaning that less and less units are required to achieve the same reconstruction error.} 
\label{cifar_reconstruction_errors_progressive}
\end{figure}

\begin{figure}[H]
\centering
\includegraphics[width=0.75\textwidth]{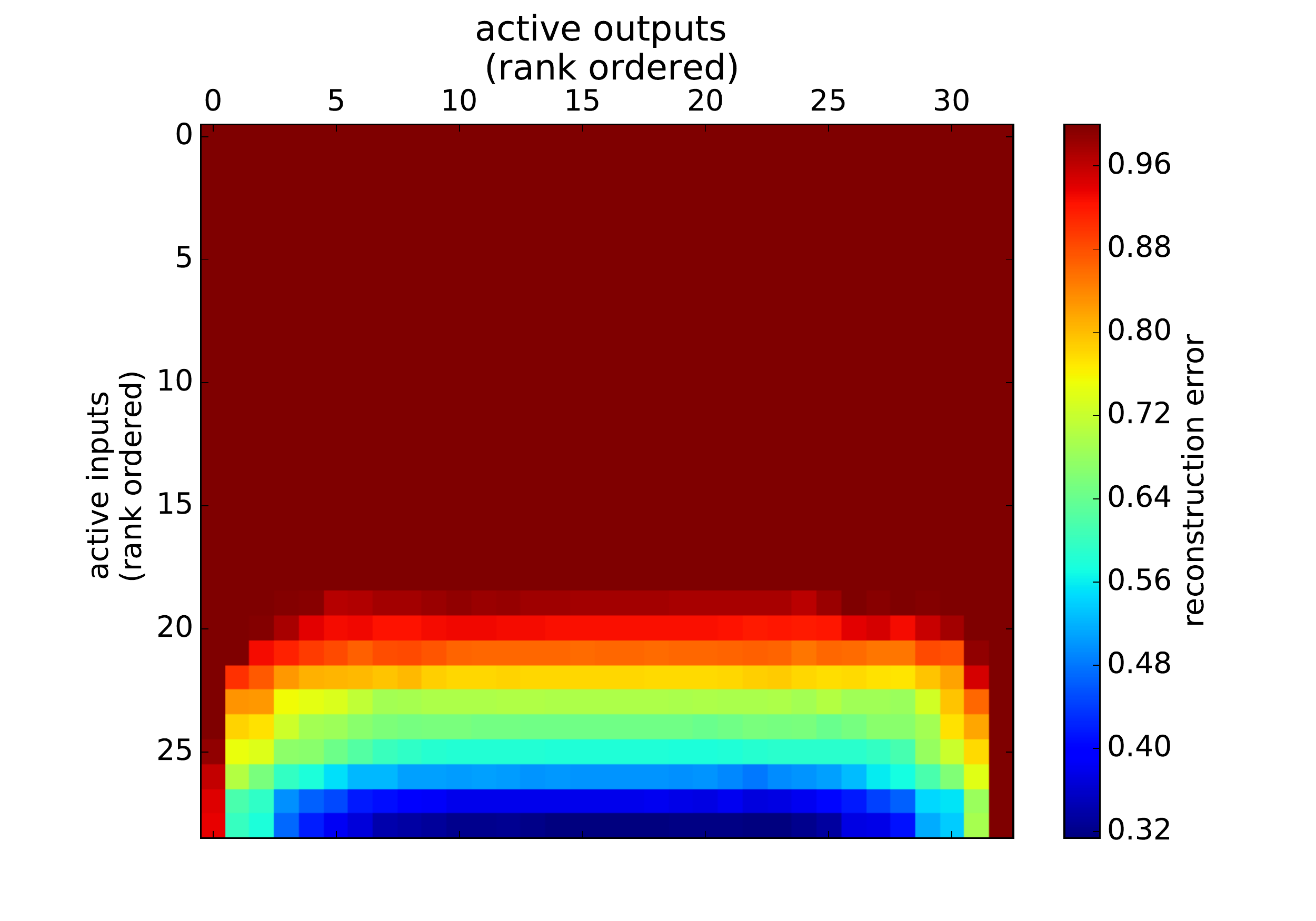}
\caption{Example of the actual learned reconstruction error surface (thresholded at 1) for the input shown in figure \ref{cifar_reconstruction_progressive}. Note the similarity with figure \ref{reconstructionsurface}, the absolute values of the raw output is taken for reconstruction to get an approximate idea of what the effect of negative output filters is on the reconstruction error.} 
\label{cifar_real_error_surface}
\end{figure}

\begin{figure}[H]
\centering
\includegraphics[width=1\textwidth]{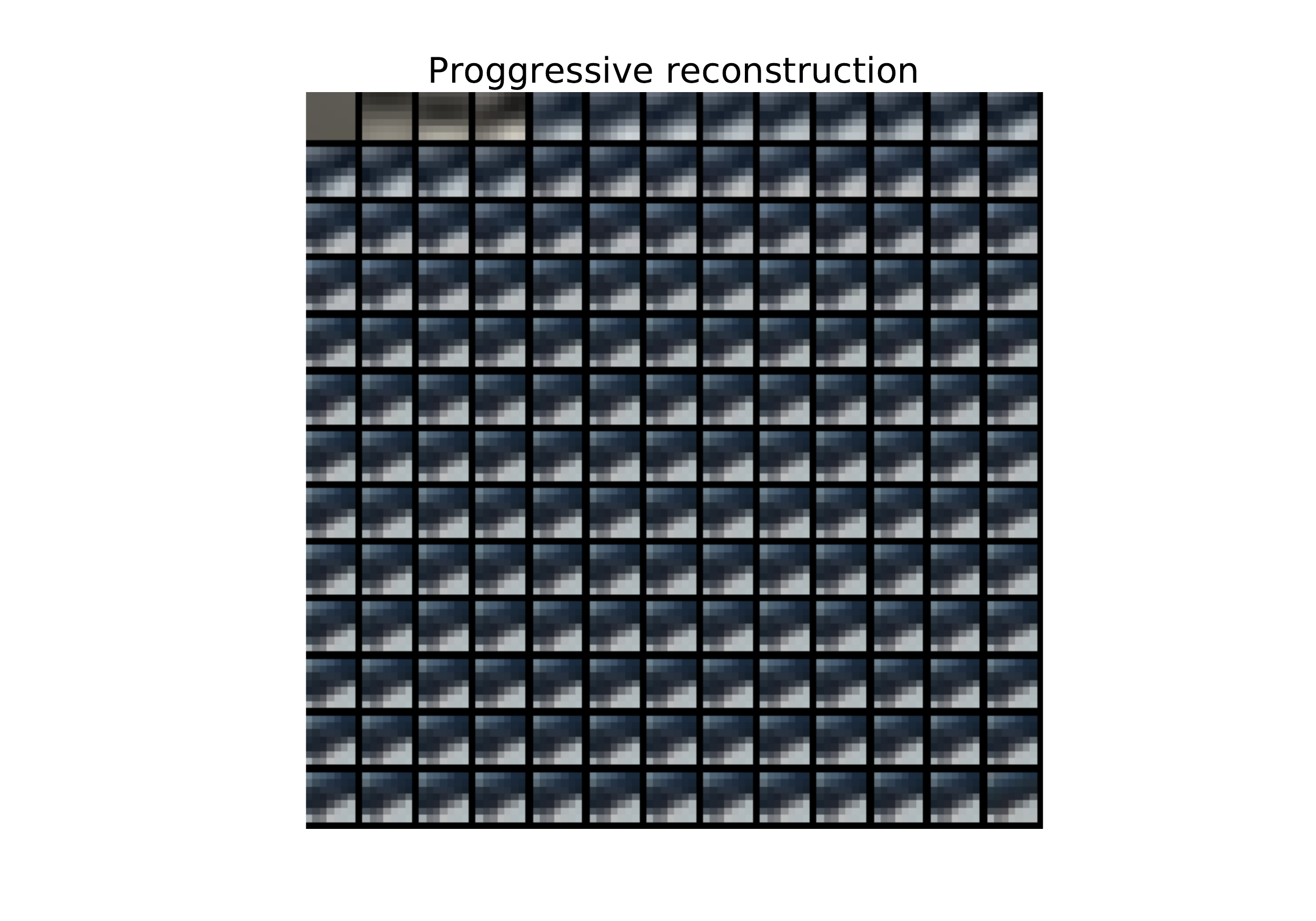}
\caption{Example of the rank ordered progressive reconstruction, the last filter (lower right corner) is overwritten with the actual current input. Each square represents the reconstruction using an additional output unit showing rapid convergence of toward the actual input using very few output units (only few units are actually active in the reconstruction, see also Figure \ref{cifar_histogram_sparsity}).} 
\label{cifar_reconstruction_progressive}
\end{figure}

\begin{figure}[H]
\centering
\includegraphics[width=0.49\textwidth, height=170pt]{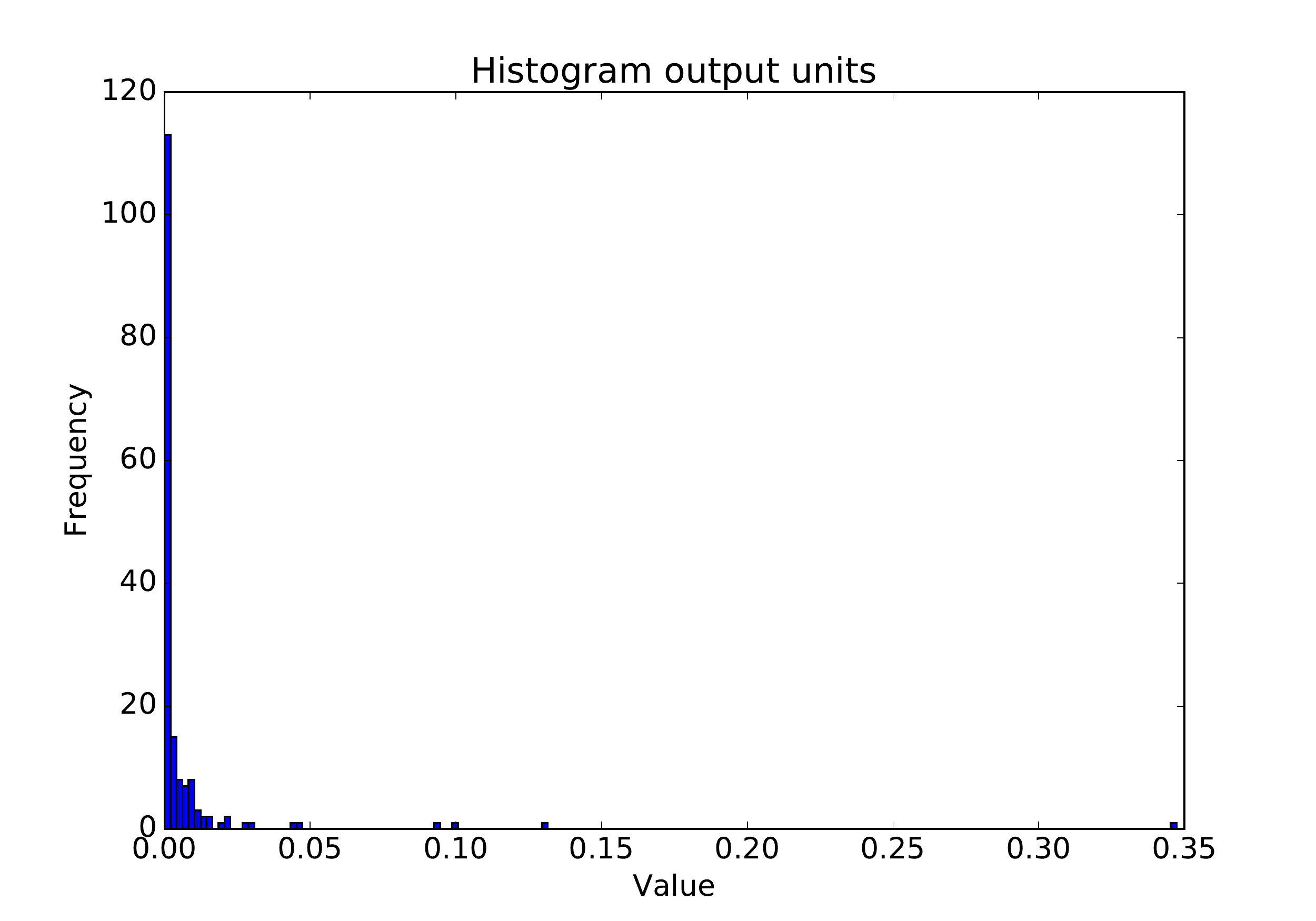}
\includegraphics[width=0.49\textwidth, height=170pt]{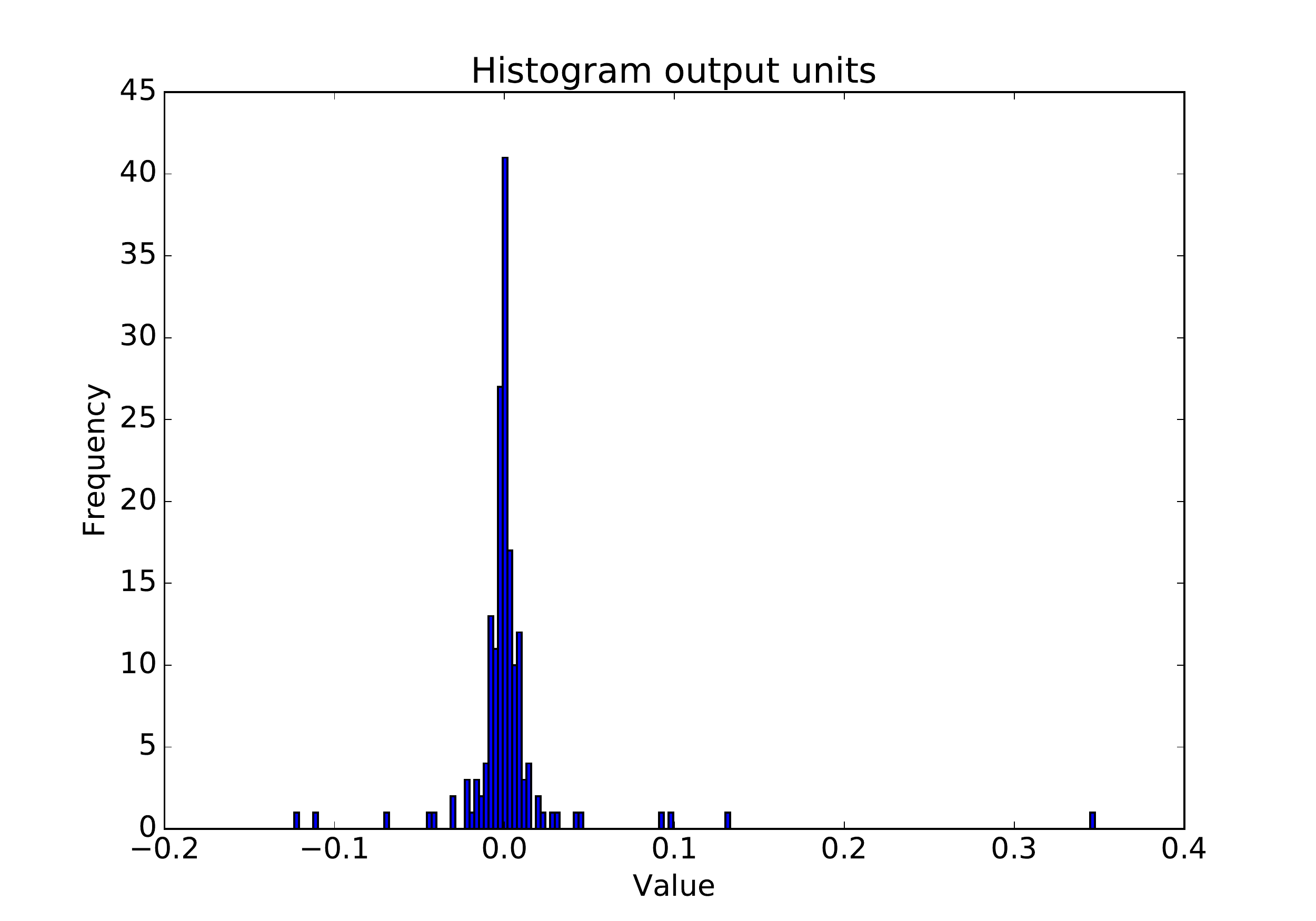}
\caption{Example of the output distribution for both the thresholded output (left) and raw output (right) for the input shown in Figure \ref{cifar_reconstruction_progressive}, showing extremely high sparsity and a sharp distribution centered around zero.} 
\label{cifar_histogram_sparsity}
\end{figure}

\subsection{Discussion} 
We can see in figure \ref{cifar_learned_filters} that we successfully learn meaningful edge and color filters over the data that are similar to previous approaches\cite{Carvalho2013, Coates2011}. Notably though it is impossible to learn redundant features due to minimizing the progressive reconstruction. However there does appear to be an issue learning some of the lower ranked units that are rarely active which could potentially be due to numerical instability, i.e. the error becomes so small that we can no longer accurately compute the gradient, or merely because they occur so rarely. It is possible that training for more epochs will eventually stabilize these filters. In figure \ref{cifar_reconstruction_errors} we see the expected extreme robustness to overfitting, the test error is nearly identical to the training error this is likely as mentioned before due to the learning of a representation that minimizes the number of active units required for reconstruction. Figure \ref{cifar_reconstruction_errors_progressive} and \ref{cifar_reconstruction_progressive} show that we can still achieve an extremely low reconstruction error with only few outputs, making the approach robust to a type of noise where lower ranked units are likely to be missing. It also allows for rapid inference using only the top $k$ number of units. This means we have indeed learned a type of rank order code over the input. If we look at figure \ref{cifar_real_error_surface} we can observe that the derived hypothetical error surface (figure \ref{reconstructionsurface}) indeed matches the real one. It additionally shows robustness to dropped inputs, even if some of the lower ranked inputs are dropped we can still reconstruct with a low error. Finally we obtain an extremely sparse output distribution (figure \ref{cifar_histogram_sparsity}) that is peaked at 0 and has a very low variance. This type of distribution is commonly explicitly applied to learn sparse distributions, instead we have implicitly learned the distribution by minimizing eq.\ref{eq2}.
			
\section{Conclusion} 			
Looking at all the results it is clear we have achieved the desired objective of minimizing the ordered $L_0$-norm on the output in eq. \ref{eq3} by minimizing eq. \ref{eq2} without setting the actual sparsity parameter $\lambda$, instead we have learned this parameter implicitly. This shows that the approach of progressive reconstruction by the sorted output is an effective way to learn sparse representations without requiring a sparsity hyperparamter. It is also easiliy parallelizable making it practical to use in a real world setting. The shown robustness to overfitting additionally allows for an easier way of online learning (streaming data), there is only one learning parameter which is the learning rate which can potentially be modified to also work in an online setting by for example making it dependent on the reconstruction error. We did observe some noisy low output values (figure \ref{cifar_histogram_sparsity}) which maybe makes it desirable to have a different type of activation function\cite{Agostinelli2015} that suppresses this noise and also normalizes high output values. One possible function for this could be a generalized logistic function, that has differentiable shape parameters. It would be interesting as well to see the networks performance on only ranking values instead of the actual output values. If this proves to be an easier task it could then be usefull to learn robustness to contrast variation by using an additional network that tries to learn the actual ranking values from the output values. Another possible extension is to apply a reverse rank ordered autoencoder on the input that learns to reconstruct the output progressively instead of the input (as mentioned in figure \ref{reconstructionsurface}) this type of reconstruction could increase robustness to a type of input noise where lower ranked inputs are more likely to be missing. One of the remaining issues is that the error at the higher ranked units is much higher which makes learning more unstable, we used norm clipping to prevent some of this instability but this is not an optimal solution. In order for low ranked units to properly learn, the units that reconstruct before it have to be stable. This could potentially be achieved by some type of reinforcement on the weights, that strengthen weights as learning progress as a function of the output. This reinforcement value could then be multiplied by the gradient to reduce the gradient updating. Finally there are the more obvious extensions, which include stacking multiple rank ordered autoencoders to learn hierarchical representations\cite{Vincent2010}, adding noise to the input and output to increase robustness and allow for denoising and generating samples\cite{Bengio2013}. A GPU implementation using minibatches would increase computational efficiency and we could classify with an extra classification layer or by reconstructing the labels progressively.

\bibliography{mybib}{}
\bibliographystyle{plain}


\end{document}